\documentclass{llncs}
\usepackage{cite}
\usepackage{amssymb}
\usepackage{amsmath}
\usepackage{color,hyperref}
\definecolor{darkblue}{rgb}{0.0,0.0,0.3}
\hypersetup{colorlinks,breaklinks,
            linkcolor=darkblue,urlcolor=darkblue,
            anchorcolor=darkblue,citecolor=darkblue}

\usepackage{url}
\usepackage{graphicx}

\newcommand{\SgpDec}{\textsc{SgpDec}}
\newcommand{\GAP}{\textsc{Gap}}

\title{Computational Understanding and Manipulation of Symmetries}

\author{Attila Egri-Nagy\inst{1,2} \and  Chrystopher L. Nehaniv\inst{1}}
\institute{
Centre for Computer Science and Informatics Research\\
University of Hertfordshire \\
Hatfield, Herts AL10 9AB, United Kingdom\\
\url{C.L.Nehaniv@herts.ac.uk}
\and
Centre for Research in Mathematics\\School of Computing, Engineering and Mathematics,\\ University of Western Sydney (Parramatta Campus)\\ Locked Bag 1797, Penrith, NSW 2751\\
\url{A.Egri-Nagy@uws.edu.au}}

\newcommand{\todo}[1]{}

\begin{document}

\maketitle

\begin{abstract}
For natural and artificial systems with some symmetry structure, computational understanding and manipulation can be achieved without learning by exploiting the algebraic structure. Here we describe this algebraic coordinatization method and apply it to permutation puzzles. Coordinatization yields a structural understanding, not just solutions for the puzzles. 
\end{abstract}

\section{Introduction}
Symmetry structure in natural and artificial systems, such as crystallography, 
chemistry, physics, and permutation puzzles, etc., can facilitate understanding and 
manipulation of these systems.
This is well-known in the mathematical sciences and from the algebraic theory of groups. 
However, until now, computational algebraic methods have not been fully exploited in Artificial Intelligence (AI).  

We show how AI systems can make use of such mathematical symmetry structure to automatically generate and manipulate hierarchical coordinate systems for finite systems whose generating symmetries are given.  Such hierarchical coordinate systems 
correspond to subgroup chains in the group structure determined by generating symmetries of the system. We demonstrate how, for any finite-state symmetry system, these coordinate systems
\begin{enumerate} 
\item can be generated automatically, i.e. deriving formal models for understanding the finite-state symmetry system, and 
\item can be deployed in manipulating the system automatically.
\end{enumerate} 
Thus, without learning,  manipulation of such systems is reduced via algebra to sequential computation in a hierarchy of simple (or simpler) group coordinates.

This general, implemented method is illustrated with examples from coordinate systems on permutation puzzles such as Rubik's cube. 

Algebra and the theory of permutation groups is well-known from 
applications of groups in chemistry and crystallography \cite{chemistrybook1},
in physics \cite{groupsPhysics}, and more recently of semigroups and groups in  systems biology, genetic regulatory networks, and biochemistry \cite{BioSys_BioChem2008,wildbook}.
Unlike machine learning or optimization techniques,
\emph{algebraic} machine intelligence can without any learning derive coordinate systems not  only on states of a structure, but also on its transformations,  i.e.\ on operations  for manipulating the structure.
These are unlike the methods,  spatial structures,  or cognitive and semantic maps presently exploited for theorem proving, path-planning, automated reasoning, but similar to coordinate system occurring elsewhere in science. 
Coordinate systems, like the ones studied here on general symmetry structures from the viewpoint of
AI applications using computer algebra, also arise in conservation laws in physics. 
As Emmy Noether showed
in the first part of the 20th century,  invariants preserved under conservation laws correspond exactly
to group-theoretic symmetry structures in physics \cite{Olver}; moreover, such invariants for physical systems 
give rise to coordinate systems of exactly the type described here \cite{clnunderstanding,wildbook}.

\subsection{Algebra, Cognitive Modeling, Coordinate Systems} 
AI techniques for cognitive modeling, machine learning and optimization so far have made relatively little use of abstract algebra.  Cognitive architectures such as SOAR,  ACT-R and SAC have been applied to build AI systems that model human cognitive capacities more or less intending to emulate faithfully the structure of cognition in humans, with applications ranging from autonomous control of aircraft based on subsymbolic rule-extraction, to human-style learning of arithmetic or 
natural language, to predictive evaluation of user interfaces (e.g.\ \cite{SOAR,SAC,postcompletion2005}). But aspects of human and machine cognition 
can also involve understanding of hierarchical processes with dynamical structure as
evidenced by the object-oriented methods \cite{NehanivCT97} or the place-value representation in human number systems \cite{ZhangNorman1995}, which have close connections to
algebra.  Here we are interested in the study of AI models that can derive, represent, and manipulate this type of knowledge, but without necessarily seeking to model human capacities faithfully.   
Models for understanding of  finite-state (and more general) dynamical systems  phenomena in general exhibit  such  feedforward, coarse-to-fine, hierarchical structure 
related to algebraic coordinate systems  
\cite{primedecomp68,NehanivCT97,wildbook}.
Our work shows how such coordinate systems can be derived and exploited automatically.

By a \emph{coordinate system} on a symmetry structure (or more general structure),  we mean a notational system in the broadest possible sense, with which a human or artificial agent can address building blocks of the structure and their relations in a decomposition, thus gaining a convenient way for grasping the structure of the original phenomenon and possibly getting tools for manipulating the components. 

An obvious example is the Descartes coordinate system, where we can uniquely specify any point of the $n$-dimensional space by $n$ coordinates. However, this is an example of a spatial and inherently non-hierarchical coordinate system for a totally homogeneous space. In general, different coordinates have different roles, addressing `parts' of the system different in size, function, etc. The natural example of a hierarchical coordinate system is our decimal positional number notation system: different coordinates correspond to different magnitudes. The examples also show that a coordinate system is very much the same thing as a decomposition: the space is decomposed into dimensions, an integer is decomposed into ones, tens, hundreds, etc. 

From this viewpoint coordinate systems become cognitive models, means of knowledge representation. The above examples show the usefulness of these coordinates, but how can we obtain these models? The good news, and the main promise of this research direction, is that we can get them automatically! In algebraic automata theory, the Prime Decomposition Theorem says that every finite state automata can be decomposed into a hierarchical coordinate system \cite{primedecomp68,eilenberg,holcombe_textbook}. Therefore the way of representing knowledge becomes algebraic, semigroup- and group-theoretical, which is really very different from other, well-established AI methods mainly based on logic (e.g.\ \cite{NewellUnifiedTheories,KnowledgeRep}).

Here we concentrate on coordinatizing symmetry structures (called `permutation groups' in algebra) via the Frobenius-Lagrange embedding. 
Related work \cite{ciaa2004poster,sgpdec,BioSys_BioChem2008} has computationally implemented the automated generation of transformation semigroup decompositions  along the lines of the Krohn-Rhodes theorem, but did not pursue the `simpler' problem of obtaining coordinate systems on permutation groups as we do here, and hence our work is both complementary and necessary for providing a full decomposition in the more general setting where operations need not be invertible.  

The use of algebra in cognitive modeling also occurs in other non-traditional applications. For
example,  the approach by Fauconnier-Turner-Goguen to conceptual blending and metaphor\cite{WayWeThink},  uses category-theoretic pushout computations with  computer algebraic implementations by Goguen to automatically
generate conceptual blends and metaphors relating two (or more) conceptual domains (knowledge of which is modeled by small categories), as well as applications to the semiotics of user-interface design, and formal specification for imperative programs built up and verified in a hierarchical manner 
\cite{Goguen1562,GoguenOBJSE}.

As there are many different ways of understanding of the same thing, there are many different coordinate systems for the same structure. Some of them may be intuitive for humans, while others will suit computational manipulation better, so the range of intelligent `users' of the coordinate system is not restricted.

\todo{The computational aspect}
Although much of the mathematical theory required here is very old, the proper computational tools were missing, therefore the idea of hierarchical coordinate systems giving understanding computationally has not been much studied nor applied. Now such tools are available open-source \cite{ciaa2004poster,sgpdec,SgpDec2014} for transformation semigroups, complemented by our work reported here for groups giving fine detail on coordinatizing the groups involved.
The mathematical significance of these coordinate systems is immediate, but as they capture one of the basic aspects of our cognitive capabilities, namely, \emph{hierarchical representation},  they might also play significant role in AI. 

\section{Mathematical Preliminaries}

Here we briefly review basic group theory and a hierarchical composition method of permutation groups.

\subsection{Essentials of Group Theory}
 A function $p: X\rightarrow X$ on the set $X$ is called a \emph{permutation} if it is one-to-one and onto (invertible).
 A \emph{permutation group} $(X,G)$ is a set $G$ of permutations closed under composition (multiplication, usually denoted by~$\cdot$), together with the state set $X$ on which the mappings act. 
It is called a \emph{symmetry group} if certain structure on $X$ is preserved by all $p\in G$. 
If it is clear form the context we omit the state set and write simply $G$. 
For $x\in X$ and $g\in G$, we write $x\cdot g$ for the result of applying permutation $g$ to $x$. 
The action is \emph{faithful} if whenever  $x \cdot g = x \cdot g'$ holds for all $x\in X$ then $g=g'$. 
The group contains the \emph{identity} $1$ and \emph{inverse} map $p^{-1}$ for each element $p$, thus everything can undone within a group. 
A group $G$ acting on $X$ by permutations need not necessarily be faithful, i.e.\ need not be a permutation group. 
The group consisting of all permutations of $n$ points is called the \emph{symmetric group} $S_n$, while $C_n$ denotes the {\em cyclic group} on $n$ objects permuted cyclically.  A subset of a group called a \emph{subgroup} if it is closed under inversion and the  group's multiplication.
If $(X,G)$ is faithful, it is a naturally a subgroup of the symmetric group on $X$.  
For more on elementary group theory see for instance \cite{hallgroup59,RobinsonGroups}, and on permutation groups see \cite{DixonMortimerPermGroups96,CameronPermGroups99}. As is standard, we use cycle notation to denote permutation group elements, e.g.\ (1 3)(2 4 5) denotes the permutation swapping $1$ and $3$, cyclically taking $2$ to $4$, $4$ to $5$, and $5$ to $2$, while leaving any other objects fixed. 

 As a special case, a group can act on itself, i.e.\ the group elements are the states and each $g_0 \in G$  maps  $G\rightarrow G$ by right multiplication, $g \mapsto g \cdot g_0$. 
This is called the \emph{right regular representation}, and it enables us to identify the group element with its effect, which will be really handy for permutation puzzles. However, this representation poses problems computationally due to the possibly large size of the state set.

\subsection{Cascade Product of Permutation Groups}

Given permutation groups $(X,G)$ and $(Y,H)$ their \emph{wreath product} is the permutation group  
$$(X,G)\wr(Y,H) \cong (X\times Y, G\rtimes D)$$ 
\noindent where $D=H^X$ is the set of all possible functions from $X$ to $H$. A state in the wreath product is expressed by two coordinates $(x,y)$, $x\in X,y\in Y$. The group elements are coordinatized similarly by $(g,d)$ where $g\in G$ and  $d:X\rightarrow H$ is the \emph{dependency function}, a `recipe' to find an element that should be applied on the second level based on the (previous) state of the first (top) level. Thus a permutation of $X\times Y$ is given by: 
$$ (x,y) \cdot (g,d) =  (x\cdot g, y\cdot d(x))$$
\noindent from which we can see that on the top level the action is independent from the bottom level, but not the other way around, hence the hierarchical nature of the wreath product. Wreath products are generally huge structures, but in practice we deal with some substructures with the dependency functions limited. The wreath product is easy to generalize for more levels.   
\section{Lagrange Coordinatizations of Groups}

The basic idea of the Lagrange Decomposition is that given a subgroup $H$ of $G$, we form the set of \emph{cosets} $G/H =\{Hg: g\in G\}$, i.e.\ the subgroup $H$ and its translates within $G$; these partition $G$, and $G$ acts on $G/H$ by right translations: $Hg \stackrel{\cdot g_0}{\mapsto} Hgg_0$. Moreover, we do not need to act on whole cosets but on arbitrary but fixed representatives of cosets ($\overline{g} \in Hg = H\overline{g}$).  The action may not be faithful, so we denote $G$ made faithful by $\tilde{G}$.\footnote{Here $\tilde{G}$ is a quotient group  $G/K$, where $K$
is the {\em core} of $H$ in $G$, that is, the largest normal subgroup of
$G$ contained in $H$. Thus $K =\bigcap_{g\in G} g^{-1} H g$. 
See standard references \cite{RobinsonGroups,CameronPermGroups99}.}

\begin{theorem}[Lagrange Decomposition] 
Let $(X,G)$ be a transitive permutation group and $(X,H)$ be a subgroup of it. Then $(X,G)$ admits the following coordinatization
$$(G/H,\tilde{G})\wr(H,H)$$
\label{LD}
corresponding to the subgroup chain $G \geq H  \geq \langle 1 \rangle$.
\end{theorem}

\noindent Thus given a state $x\in X$ we will coordinatize it by $\tilde{x}=(x_1,x_2)$, where each $x_i$ is a coset representative ($x_1\in G/H, x_2\in H/\langle 1 \rangle$).

By refining the underlying subgroup chain  we can make the component 
groups much simpler or {\em simple} (i.e.\ having only trivial homomorphic images). 
This allows one to iterate the Lagrange coordinatization so the problem of understanding the permutation group $(X,G)$ is reduced to understanding much simpler permutation groups linked up in a feedforward manner.  
Therefore getting a coordinatization corresponds to devising a subgroup chain. 

For building subgroup chains certain subgroups are  very useful. The \emph{stabilizer} $G_a$ is the subgroup of $G$, which fixes $x\in X$. Point-wise and set-wise stabilizers can be defined for sets of states as well. By iterating and refining the Lagrange construction, we have

\begin{theorem}[Frobenius-Lagrange Coordinatization] 
\label{FL}
Let $(X,G)$ be a transitive permutation group and
let $G=G_1 \geq G_2 \geq \cdots \geq G_{n+1}=G_a$ be a subgroup chain ending at the stabilizer $G_a$ of some state $a\in X$. Then, in the notation above, 
$(X,G)$ is coordinatized by embedding in the wreath product $$(G_1/G_2, \tilde{G_1})\wr \cdots \wr (G_n/G_{n+1},\tilde{G_n})$$
where $\tilde{G_i}$ is $G_i$ modulo the core of $G_{i+1}$ in $G_i$. Moreover, since $(X,G)$ corresponds isomorphically to $(G/G_a,G)$, the  states have coordinates given 
by the bijection  $$G_a x_n \cdots x_1 \Leftrightarrow (x_1,\ldots,x_n)$$ where each $x_i$ is the fixed coset representative of $G_{i+1}x_i$ in $G_i$.  
\end{theorem}

\noindent{\bf Remarks.}
(1) The number of coordinate tuples  in Theorem~\ref{FL} is exactly  $|X|$. Each state has a unique coordinatization.
(2) In Theorem~\ref{LD} the number of possible coordinate tuples is $|G|=|X|\cdot |G_a|$, for any $a\in X$,
as each point has exactly $|G_a|$ different possible coordinatizations.
\section{Coordinate Manipulation}

The Frobenius-Lagrange decomposition gives us, in terms of a coordinate system, a structured view of the group, i.e.\ we can address its parts conveniently and with arbitrary precision. However, we would like to use the coordinate system dynamically, not just the statical view. We would like to calculate with it, finding manipulative operations taking one state to another desired state, or, equivalently, from one tuple of coordinate values to another one using the elementary symmetry operations of the original structure.

\subsection{Component Actions}

For establishing the connection between the original group and the coordinatized one we need to have a way to express a permutation as coordinate actions. Given a group element $g\in G$ and a coordinatized state $\tilde{x}=(x_1,\ldots, x_n)$, we can calculate the coordinatewise component actions by the following recursive calculations:
$${g}_1 := g $$ 
$${g}_{i+1} := x_i \cdot {g}_i \cdot \big( \overline{x_i \cdot {g}_i} \big)^{-1} \in G_{i+1}$$ 
Thus $g$ on $\tilde{x}$ is coordinatized as $\tilde{g}(\tilde{x}) = (g_1,\dots,g_n)$.  Note that  generally  $x_i \cdot {g}_i$ does not equal to $ \overline{x_i \cdot {g}_i}$, so $g_i$ is not the identity.
Note the hierarchical structure: $g_i$ depends only on $g$ and $(x_1,\ldots, x_{i-1})$. The action of $g$ in coordinatized form is then 
$$(x_1,\ldots,x_n)\cdot \tilde{g} = (x_1\cdot g_1, \ldots , x_n \cdot g_n).$$

\subsubsection{Killing and Building by Levels}
We call the coordinate tuple the \emph{base state}, if it consists of only the identities (as coset representatives), which clearly represents the identity of the original group. Given an arbitrary coordinatized state $\tilde{x} =(x_1, \dots, x_n)$, we call the coordinatewise changes of values from $x_i$ to $1$ (top-down) \emph{`killing by levels'}. This is accomplished by simply applying the inverse of  the coset representatives, in order.
An example is shown in Fig.~\ref{gap:beattheclock}.
Conversely, \emph{`building by levels'} is accomplished bottom-up by successively applying the coset representatives, i.e.\ elementary generating symmetries whose product is the given coset representative $x_i$ for the $i$th component, in the order  $x_n$ then  $x_{n-1}$ ,..., and finally  $x_1$  to move from the `solved state' $(1,\ldots,1)$
 to create state $\tilde{x}$ bottom-up. Moreover, one can compute elements that change only a single coordinate to a desired value.

\subsubsection{Global Transformation via Coordinate Values}
Since we work with groups, whenever we make any action, it can be undone by an inverse, thus reversing the killing by levels we can go from the base state to any other coordinate value combination. Thus going from $\tilde{x}$ to $\tilde{y}$ coordinatewise can be achieved simply combining the level-killers of $\tilde{x}$ with the level-builders of $\tilde{y}$. More efficient solutions are generally possible (and implemented), but this provides at least one way to do it using the hierarchical coordinate system.

\section{An Application: Permutation Puzzles}

\begin{figure}[t]
\small
{\bf
  \begin{verbatim}
gap> pocket_cube_F := (9,10,11,12)(4,13,22,7)(3,16,21,6);;
gap> pocket_cube_R := (13,14,15,16)(2,20,22,10)(3,17,23,11);;
gap> pocket_cube_U := (1,2,3,4)(5,17,13,9)(6,18,14,10);;
gap> pocket_cube_L := (5,6,7,8)(1,9,21,19)(4,12,24,18);;
gap> pocket_cube_D := (21,22,23,24)(12,16,20,8)(11,15,19,7);;
gap> pocket_cube_B := (17,18,19,20)(1,8,23,14)(2,5,24,15);;
gap> pocket_cube_gens := [pocket_cube_U, pocket_cube_L, pocket_cube_F,
>                     pocket_cube_R, pocket_cube_B, pocket_cube_D];;
gap> pocket_cube_gen_names := ["U","L","F","R","B","D"];;
gap> pocket_cube := GroupByGenerators(pocket_cube_gens);
<permutation group with 6 generators>
gap> scrambled := Random(pocket_cube);
(1,19,20,3,18,24,15,10,5,8,23,13)(2,6,7,17,4,12,14,9,21)
gap> inverse := Inverse(scrambled);
(1,13,23,8,5,10,15,24,18,3,20,19)(2,21,9,14,12,4,17,7,6)
gap> epi := EpimorphismFromFreeGroup(pocket_cube:names:=
                                ["U","L","F","R","B","D"]);
gap> sequence := PreImagesRepresentative(epi,inverse);                             
U^-1*R*U^-1*F*L*D^-1*F*L^-1*U*F^-1*U^-1*F*U*L^-1*F^-1*L*F*L*
U^-1*F^-1*U^-1*L*F*L^-1*F^-1*U^-1*F*U*R^-1*U*F^-2*U^-2*L
gap> Length(sequence);
35
\end{verbatim}
}
\normalsize
\caption{{\bf Getting a solution for the Pocket Cube.} This excerpt from a \GAP~interactive session.
The possible moves as generators (basic symmetry operations $F$, $R$, $U$, $L$, $D$, $B$ rotating a face 90$^\circ$)  for the Pocket Cube are defined together with names. 
Group elements are represented here in cycle notation, with $()$ denoting the identity element (or, according to the right regular representation, the solved state). 
For obtaining a solution we simply express the inverse of permutation representing the scrambled state as a sequence of generators.}
\label{gap:simplesolution}
\end{figure}

Permutation puzzles are one person games where the moves are permutations, elements of a group \cite{Joyner}.  Natural problems for such puzzles are: 
\begin{enumerate}
 \item How can one go, via elementary legal moves, from one configuration $x$ of the puzzle to a standard ``solved'' configuration?
\item More generally, how can one go from configuration $x$ to another arbitrarily selected configuration $y$?
\end{enumerate}


The main quest of permutation puzzles is often to find a shortest sequence of moves that leads to the solution (bounded by the diameter of the Cayley graph of the underlying group). Though the idea of of nested coset space actions is also used in tackling the shortest solution problem (e.g.~Thistlethwaite's Algorithm), it is not our aim here.
We would like to facilitate understanding, but usually the quickest solutions are `dirty tricks' that are very difficult to grasp and one often has to fall back on simple memorizing. 
In fact, nonoptimal solutions are easy to get (see Fig.\ \ref{gap:simplesolution}). One can follow the steps to solve the cube without gaining any understanding.
\emph{ Can we learn to solve the cube from the above answer?}
\emph{ Can we identify, talk about and solve subproblems?}
\emph{ Can we devise and compare different solving strategies?}
We aim to answer these questions.

\subsection{Coordinatizing Rubik's Cubes}

The $3\times 3\times 3$ Rubik's Cube is probably the most popular permutation puzzle.

What does it mean `to know the Rubik's cube'? The question usually boils down to the ability to solve the cube. By asking a cube-fan he/she would give a few tricks, recipes to apply in certain situations. By learning these algorithms one can learn to solve the cube, but does it imply understanding the cube, i.e.\ grasping how certain sequences of moves work and seeing why they work? Not necessarily.

We claim that understanding comes with imposing a coordinate system on the underlying algebraic structure. Here we demonstrate this on the symmetry group of the Pocket Cube, which is the $2\times 2\times 2$ version of the Rubik's Cube. The moves are the 90 degree clockwise rotations of the 6 sides. By the Frobenius-Lagrange decomposition we know that each coordinate system corresponds to a subgroup chain, so devising new algorithms for solving the cube is equivalent to constructing subgroup chains.
Such a coordinate system encodes a `global viewpoint' in which one solves by successive approximation, with manipulations going from coarse to fine resolution, and converging in terms of moving from  natural, abstract states to fully specified states. The group of the cube acts on the set of configurations in such a way that any non-trivial permutation yields a different result on the `solved state', thus the stabilizer of this state is $\langle 1 \rangle$, so each group element corresponds to a unique configuration. Hence by the remark following Theorem~\ref{FL}, for
any coordinate system arising from  any subgroup chain down to $\langle 1 \rangle$,  
 each configuration of the cube has a unique coordinatized form.
Examples derived computationally follow (see also Fig.~\ref{gap:beattheclock}).

\subsubsection{Pocket Cube: Cornerwise Decomposition}
One can solve the cube in a rather long, systematic step-by-step fashion: get the position and then the orientation of the first corner right, then proceed to the next corner until the cube is solved. In the subgroup chain we put the stabilizer of the position of a corner, then continue with the stabilizer of the orientation of the corner within the position stabilizing subgroup. Then we repeat the whole process for another corner. The chain yields the following coordinatization:  
$$S_8\wr C_3 \wr S_7\wr C_3 \wr S_6\wr C_3 \wr S_5\wr C_3 \wr S_4\wr C_3 \wr S_3\wr C_3 \wr C_2\wr C_3$$
where the top level component  $S_8$ acts on 8 states/coordinate values, representing the 8 possible positions of the first stabilized corner. Therefore killing the first level will put the corner in the right position. The coordinate values on the second level correspond the 3 possible rotational states of the corner.  The 3rd and 4th level similarly encode the second corner, and so on. 
\begin{figure}
\begin{center}
\includegraphics{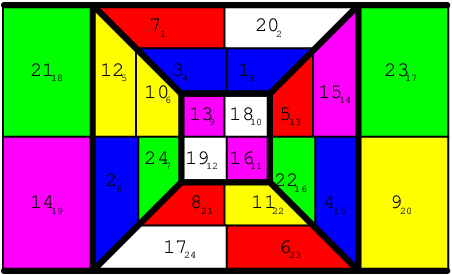}
\end{center}
\caption{A random, `scrambled' configuration of the Pocket Cube, coordinatized by the cornerwise decomposition as $( 8, 2, 5, 3, 2, 2, 5, 1, 2, 3, 3, 3, 1, 2 )$. Note: coset
representatives have been  integer-encoded in these examples.}
\label{scrambled}
\end{figure}

\begin{figure}
\begin{center}
\includegraphics{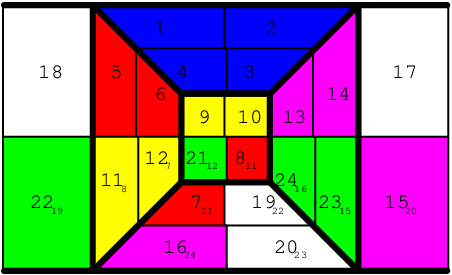}
\end{center}
\caption{Pocket Cube configuration after killing the top 9  levels out of 14. 
$(1,1,1,1,1,1,1,1,1,3,3,1,2)$}
\label{halfsolved}
\end{figure}

\begin{figure}
\begin{center}
\includegraphics{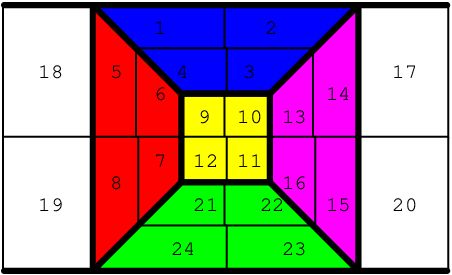}
\end{center}
\caption{The solved state of the Pocket Cube with coordinates $(1,1,1,1,1,1,1,1,1,1,1,1,1,1)$}
\label{solved}
\end{figure}

\subsubsection{Another Model for Understanding: Permute the Corners, then `Beat the Clock'}
\begin{figure}
\small
{\bf
  \begin{verbatim}
gap> #creating a subgroup chain from the chief series
gap> subgroupchain := ShallowCopy(ChiefSeries(pocket_cube));;
gap> Remove(subgroupchain,2);;
gap> #getting the hierarchical components
gap> cags := CosetActionGroups(subgroupchain);;
gap> StructureDescription(cags.components[1]);
"S8"
gap> StructureDescription(cags.components[2]);
"C3 x C3 x C3 x C3 x C3 x C3 x C3"
gap> #solving the cube from a random state
gap> scrambled := Random(pocket_cube);
(1,10,12,6,23,14,16,24)(2,22,19,5,3,21,4,15)(7,9,20,17,11,8,18,13)
gap> coordinates := Perm2Coords(scrambled, cags.transversals);
[ 22578, 552 ]
gap> levelkillers := LevelKillers(coordinates,cags.transversals);
[ (1,19,22,2,15,9,7,3)(4,21,10,18,24,16,14,23)(5,8,11,17,20,6,12,13),
  (1,5,18)(3,13,10)(4,6,9)(8,19,24) ]
gap> halfsolved := scrambled * levelkillers[1];
(1,18,5)(3,10,13)(4,9,6)(8,24,19)
gap> halfsolvedcoords := Perm2Coords(halfsolved,cags.transversals);
[ 1, 552 ]
gap> halfsolved * levelkillers[2] = ();
true
\end{verbatim}
}
\normalsize
\caption{{\bf Deriving a Coordinate System for Pocket Cube and Solving via Killing by Levels.} 
The subgroup chain for the decomposition is the modified chief series of the group. Then the decomposition is calculated yielding the two level coordinatization. Then a scrambled  (random element of the Pocket Cube permutation group) is shown in coordinate format. Finally, the scrambled cube is solved by levels of this hierarchical coordinate system, top-down, using level-killers (see text), which are also expressed as an [unoptimized] sequence of the original generators. }
\label{gap:beattheclock}
\end{figure}

Contrasting to the previous, very machine-minded solution, here is another one which is short, and reveals the existence of a different puzzle within the Pocket Cube:
$$S_8 \wr \prod_{i=1}^{7}C_3.$$
The top level component is the right regular representation of the now familiar symmetric group permuting the 8 corners. The second level is the direct product of 7 copies of modulo 3 counters (the orientation group of corners). It is to be noted that there are  not 8 copies, otherwise every corner could be rotated independently from the other corners (and that would be rather easy to solve). Actually solving the bottom level is the same type of problem as the Rubik's Clock \cite{RubiksClock}, which is an array of connected modulo 12 counters. As the underlying group is commutative, it is easier to solve since the order of operations generating 
this subpuzzle does not matter in this lowest level. 

For an example computational session using our decomposition package \SgpDec \cite{sgpdec,SgpDec2014} in \GAP \cite{GAP4} see Fig.~\ref{gap:beattheclock}.

\subsubsection{3 $\times$ 3 $\times$ 3 Rubik's Cube}
Going to the standard Cube we immediately meet some difficulty, as its group is not a transitive one. Therefore,  using $(G,G)$ we can get a decomposition which solves the corners as in the Pocket Cube  and nearly separately and in parallel the remaining non-corner  middle faces (those not at the corners, not in the middle of a side) on which cube group is transitive. Then we can proceed by coordinatizing and solving the Pocket Cube and this middle cube puzzles independently.

\section{Conclusion and Future Work}

We have shown how different subgroup chains in a permutation group correspond to different Frobenius-Lagrange coordinatizations of that permutation group, as well as to different solving strategies for manipulation. 
Coordinatewise manipulation of the permutation group via short or minimal length words over group's basic generators is an 
easily achieved next step.  In particular, we showed that solving strategies for permutation puzzles can be represented by a subgroup chain, which determines a hierarchical decomposition.

For exploitation of the idea of hierarchical coordinatization in more general settings, groups can generalized to semigroups in order allow the possibility of some irreversible manipulations \cite{sgpdec,SgpDec2014,wildbook}.

\bibliographystyle{plain}
\bibliography{../full.bib}

\begin{thebibliography}{10}

\bibitem{KnowledgeRep}
Ronald~J. Brachman and Hector~J. Levesque.
\newblock {\em Knowledge Representation and Reasoning}.
\newblock Morgan Kaufmann, 2004.

\bibitem{CameronPermGroups99}
Peter~J. Cameron.
\newblock {\em Permutation Groups}.
\newblock London Mathematical Society, 1999.

\bibitem{DixonMortimerPermGroups96}
John~D. Dixon and Brian Mortimer.
\newblock {\em Permutation Groups}.
\newblock Graduate Texts in Mathematics 163. Springer, 1996.

\bibitem{BioSys_BioChem2008}
A.~Egri-Nagy, C.~L. Nehaniv, J.~L. Rhodes, and M.~J. Schilstra.
\newblock Automatic analysis of computation in biochemical reactions.
\newblock {\em BioSystems}, 94(1-2):126--134, 2008.

\bibitem{SgpDec2014}
Attila Egri-Nagy, James~D. Mitchell, and Chrystopher~L. Nehaniv.
\newblock Sgpdec: Cascade (de)compositions of finite transformation semigroups
  and permutation groups.
\newblock In Hoon Hong and Chee Yap, editors, {\em Mathematical Software –
  ICMS 2014}, volume 8592 of {\em Lecture Notes in Computer Science}, pages
  75--82. Springer Berlin Heidelberg, 2014.

\bibitem{ciaa2004poster}
Attila Egri-Nagy and Chrystopher~L. Nehaniv.
\newblock Algebraic hierarchical decomposition of finite state automata:
  Comparison of implementations for {K}rohn-{R}hodes {T}heory.
\newblock In {\em Conference on Implementations and Applications of Automata
  CIAA 2004}, volume 3317 of {\em Springer Lecture Notes in Computer Science},
  pages 315--316, 2004.

\bibitem{sgpdec}
Attila Egri-Nagy, Chrystopher~L. Nehaniv, and James~D. Mitchell.
\newblock {\em \textsc{{S}gp{D}ec} -- software package for hierarchical
  decompositions and coordinate systems, Version 0.7+}, 2013.
\newblock \href{http://sgpdec.sf.net}{\url{http://sgpdec.sf.net}}.

\bibitem{eilenberg}
Samuel Eilenberg.
\newblock {\em Automata, Languages and Machines}, volume~B.
\newblock Academic Press, 1976.

\bibitem{WayWeThink}
Gilles Fauconnier and Mark Turner.
\newblock {\em The Way We Think: Conceptual Blending and the Mind's Hidden
  Complexities}.
\newblock Basic Books, 2003.

\bibitem{GAP4}
The GAP~Group.
\newblock {\em {\textsc{GAP} -- Groups, Algorithms, and Programming, Version
  4.7.5}}, 2014.
\newblock \href{http://www.gap-system.org}{\url{www.gap-system.org}}.

\bibitem{Goguen1562}
Joseph Goguen.
\newblock An introduction to algebraic semiotics with application to user
  interface design.
\newblock In {\em Computation for Metaphor, Analogy and Agents}, volume 1562 of
  Lecture Notes in Artificial Intelligence, pages 242--291. Springer Verlag,
  1999.

\bibitem{GoguenOBJSE}
Joseph~A. Goguen and Grant Malcolm, editors.
\newblock {\em Software Engineering with {O}{B}{J}: Algebraic Specification in
  Action}.
\newblock Springer Verlag, 2000.

\bibitem{hallgroup59}
Marshall Hall.
\newblock {\em The Theory of Groups}.
\newblock The Macmillan Company, New York, 1959.

\bibitem{holcombe_textbook}
W.~M.~L. Holcombe.
\newblock {\em Algebraic Automata Theory}.
\newblock Cambridge University Press, 1982.

\bibitem{groupsPhysics}
Hugh~F. Jones.
\newblock {\em Group Theory, Representations and Physics}.
\newblock Adam Hilger, 1990.

\bibitem{Joyner}
David Joyner.
\newblock {\em Adventures in Group Theory}.
\newblock John Hopkins University Press, 2002.

\bibitem{primedecomp68}
Kenneth Krohn, John~L. Rhodes, and Bret~R. Tilson.
\newblock The prime decomposition theorem of the algebraic theory of machines.
\newblock In Michael~A. Arbib, editor, {\em Algebraic Theory of Machines,
  Languages, and Semigroups}, chapter~5, pages 81--125. Academic Press, 1968.

\bibitem{SOAR}
John Laird, Allen Newell, , and Paul Rosenbloom.
\newblock {S}{O}{A}{R}: An architecture for general intelligence.
\newblock {\em Artificial Intelligence}, 33(1):1--64, September 1987.

\bibitem{postcompletion2005}
S.~Y.~W. Li, A.~Blandford, P.~Cairns, and R.~M. Young.
\newblock Post-completion errors in problem solving.
\newblock In {\em Proceedings of the Twenty-Seventh Annual Conference of the
  Cognitive Science Society}, Hillsdale, NJ, 2005. Lawrence Erlbaum Associates.

\bibitem{NehanivCT97}
C.~L. Nehaniv.
\newblock Algebraic models for understanding: Coordinate systems and cognitive
  empowerment.
\newblock In {\em Proc.\ Second International Conference on Cognitive
  Technology: Humanizing the Information Age}, pages 147--162. IEEE Computer
  Society Press, 1997.

\bibitem{clnunderstanding}
Chrystopher~L. Nehaniv.
\newblock Algebra and formal models of understanding.
\newblock In Masami Ito, editor, {\em Semigroups, Formal Languages and Computer
  Systems}, volume 960, pages 145--154. Kyoto Research Institute for
  Mathematics Sciences, RIMS Kokyuroku, August 1996.

\bibitem{NewellUnifiedTheories}
Allen Newell.
\newblock {\em Unified Theories of Cognition}.
\newblock Harvard University Press, 1990.

\bibitem{Olver}
Peter~J. Olver.
\newblock {\em Applications of Lie Groups to Differential Equations}.
\newblock Springer Verlag, 2nd edition, 2000.

\bibitem{chemistrybook1}
K.~V. Raman.
\newblock {\em Group Theory and Its Applications to Chemistry}.
\newblock Tata McGraw-Hill, 2004.

\bibitem{SAC}
L.~M. Reder and C.~D. Schunn.
\newblock Metacognition does not imply awareness: Strategy choice is governed
  by implicit learning and memory.
\newblock In Lynne~M. Reder, editor, {\em Implicit Memory and Metacognition},
  pages 45--77. Erlbaum, Hillsdale, NJ, 1996.

\bibitem{wildbook}
John Rhodes.
\newblock {\em {Applications of Automata Theory and Algebra via the
  Mathematical Theory of Complexity to Biology, Physics, Psychology,
  Philosophy, and Games}}.
\newblock World Scientific Press, 2009.
\newblock Foreword by Morris W.\ Hirsch, edited by Chrystopher L.\ Nehaniv
  (Original version: University of California at Berkeley, Mathematics Library,
  1971).

\bibitem{RobinsonGroups}
Derek J.~S. Robinson.
\newblock {\em A Course in the Theory of Groups}.
\newblock Springer, 2nd edition, 1995.

\bibitem{RubiksClock}
Christopher~C. Wiggs and Christopher~J. Taylor.
\newblock Mechanical puzzle marketed as {R}ubik's {C}lock.
\newblock Patent EP0322085, 1989.

\bibitem{ZhangNorman1995}
J.~Zhang and D.~A. Norman.
\newblock A representational analysis of numeration systems.
\newblock {\em Cognition}, 57:271--295, 1995.

\end{thebibliography}

\end{document}